\documentclass[letterpaper, 10 pt, conference]{ieeeconf}  
\IEEEoverridecommandlockouts  
\usepackage[utf8]{inputenc}
\usepackage{amsmath, amssymb}
\usepackage{accents}
\usepackage{booktabs}
\usepackage{graphicx}
\usepackage{xcolor}

\usepackage{algorithm}
\usepackage[noend]{algpseudocode}

\makeatletter
\def\BState{\State\hskip-\ALG@thistlm}
\makeatother

\title{Prioritized Kinematic Control of Joint-Constrained Head-Eye Robots using the Intermediate Value Approach}
\author{Steven Jens Jorgensen$^{\dagger+}$, Orion Campbell$^\dagger$, Travis Llado$^\dagger$, Jaemin Lee$^\dagger$, Brandon Shang$^*$ , and Luis Sentis$^{\dagger\dagger}$  
\thanks{$^\dagger$The Department of Mechanical Engineering, $^*$Electrical and Computer Engineering, and $^{\dagger\dagger}$Aerospace Engineering and Engineering Mechanics at the University of Texas at Austin.}
\thanks{$^+$This work is partially supported by a NASA Space Technology Research Fellowship Grant Number NNX15AQ42H}
\thanks{The authors are grateful to the members of the Human-Centered Robotics Lab in UT Austin for their input}
}

\begin{document}
\maketitle
\thispagestyle{empty}
\pagestyle{empty}

\begin{abstract}
Existing gaze controllers for head-eye robots can only handle single fixation points. Here, a generic controller for head-eye robots capable of executing simultaneous and prioritized fixation trajectories in Cartesian space is presented. This enables the specification of multiple operational-space behaviors with priority such that the execution of a low priority head orientation task does not disturb 
the satisfaction of a higher prioritized eye gaze task. Through our approach, the head-eye robot inherently gains the biomimetic vestibulo-ocular reflex (VOR), which is the ability of gaze stabilization under self generated movements. The described controller utilizes recursive null space projections to encode joint limit constraints and task priorities. To handle the solution discontinuity that occurs when joint limit tasks are inserted or removed as a constraint, the  Intermediate Desired Value (IDV) approach is applied. Experimental validation of the controller's properties is demonstrated with the Dreamer humanoid robot. Our contribution is on (1) the formulation of a desired gaze task as an operational space orientation task, (2) the application details of the IDV approach for the prioritized head-eye robot controller that can handle intermediate joint constraints, and (3) a minimum-jerk  specification for behavior and trajectory generation in Cartesian space.

\end{abstract}

\section{INTRODUCTION}
\begin{figure}
\centerline{\includegraphics[width=0.95\columnwidth]{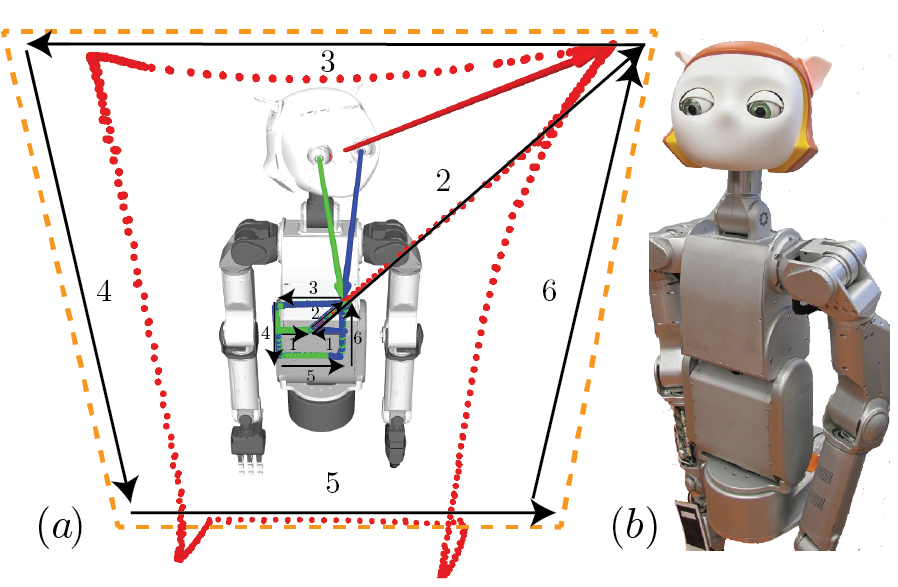}}
\caption{The Dreamer robot executing prioritized gaze tasks using the described projection-based controller, in which the eye gaze tasks have higher priority than the head gaze tasks. (a) Starting from the center, the eyes are to create a small square in a counter-clockwise direction and the head is commanded to create a big square (orange dotted-line) in a clockwise direction. The numbered arrows indicate the waypoint trajectory order. The red, blue, green spheres indicate the actual trajectory of the head, left eye, and right eye respectively. Due to joint limits, both trajectories cannot be accomplished. However, task priority and solution smoothness are preserved as joint limit tasks are automatically inserted via the Intermediate Desired Value approach. The figure shows the controller satisfying the higher prioritized eye gaze task by compromising the lower priority head gaze task. (b) Dreamer in the final configuration after executing the trajectories.
}
\label{fig:dreamer_prioritized_square}
\end{figure}
The control of a robot's gaze behavior has practical use in Human-Robot Interaction as gaze cues can be used to initiate and ensure joint attention \cite{huang2011effects}, communicate intentions and engagement \cite{moon2014meet, breazeal2005effects}, shape conversation roles \cite{mutlu2009footing}, and convey non-verbal expressions or emotions pertinent during social interactions \cite{admoni2017social,kleinke1986gaze}. The gaze behavior for anthropomorphic robots with a head-eye mechanism are even more important for human likability \cite{disalvo2002all}. Here, the definition of a gaze task is extended to any end-effector that can point towards a fixation point. Thus, in addition to each eye having its own fixation point, the robot head can also have a fixation point.


While control methods are available for specifying 3D gaze fixation tasks \cite{takanishi1997development,lopes2009biomimetic,roncone2016cartesian,drake, SentisPhd}, control formulations that can handle multiple 3D gaze fixation points with priorities for generic head-eye robots are largely lacking. The control formulation presented here addresses that need by focusing on the precise control of multiple gaze fixation points for generic head-eye robots. Concretely, the proposed controller can handle multiple gaze orientation tasks and execute the desired tasks with prioritization. Figure~\ref{fig:dreamer_prioritized_square} shows how the described controller executes three orientation tasks (two for the eyes and one for the head) with priorities under joint limits.

The prioritized controller is based on the whole-body control of robots in the operational space using null-space projection \cite{SentisPhd}, \cite{sentis2005control}. Using this formulation, control policies for any robot require merely identifying the correct Jacobians and operational task description. Thus, formulating a controller in this manner creates a generic head-eye controller. 

Null-space projection techniques are popular for prioritized control of redundant robots \cite{baerlocher1998task, sentis2004prioritized, sugiura2007real, albu2007dlr} as they are analyzable \cite{antonelli2008stability} and computationally efficient \cite{chang2000operational}. However these controllers fail to satisfy task specification without the inclusion of joint limits. Since joint limits are intermediate, the joint limit constraints need to be constantly inserted or removed from the task specification. However, doing so changes the dimension of the task Jacobian causing discontinuities when performing pseudo-inverses or optimizations \cite{keith2011analysis}. To handle this issue, the Intermediate Desired Value (IDV) \cite{lee2012intermediate, han2013robot} approach is utilized, which can automatically insert joint limit tasks and preserve solution continuity. 

The paper is organized as follows. Section II provides a discussion of related works on the control of head-eye robots. Section III describes the technical approach of (i) extracting the task Jacobian for head-eye robots, (ii) expressing the desired gaze fixation point as an operational space task, (iii) detailing the IDV-based prioritized controller, and (iv) generating minimum-jerk based Cartesian-space gaze trajectories. Section IV and V show experimental results on the Dreamer robot and provide concluding discussions.

\section{Related Works}
Due to the importance of gaze behavior, there are many approaches to implementing gaze controllers. For research applications that need immediate results, gaze control can be as simple as executing predetermined configurations to simulate gaze aversion \cite{andrist2014conversational} in conversations. Approximate gaze control can also be sufficient if the imitation of human cognition \cite{breazeal1999context}, or the study of biomimicry \cite{shibata2000biomimetic} are more important. 

For robots that need precise gaze control with biomimetic behavior, the implementation of such controllers is split between achieving gaze in a 2D image space or a 3D fixation point. Examples of the the former creates a mapping between joint positions and the optical flow of the 2D image space \cite{brooks1999cog,EdsingerPhd,vijayakumar2001overt}. For the latter, reasoning about the robot kinematics and trigonometric constraints can give a direct inverse kinematics solution \cite{takanishi1997development}, but this is restricted to similarly configured robots. Other examples of 3D-cartesian controllers capable of executing biomimetic 3D gaze fixation tasks include \cite{lopes2009biomimetic} combining human data and established state-space control methods, as well as a completely optimization-based method \cite{roncone2016cartesian} to achieve 6-DoF gaze cartesian control. However, the latter is specifically formulated for a robot with only two eyes having a single fixation point for both the head and eyes. 

Thus, all the gaze control formulations above are not general enough for generic head-eye robots in that it cannot handle multiple fixation points and that task priority is non-existent. A brute-force method is also available via nonlinear optimization with the Drake control tool box \cite{drake}, which can specify a single gaze task as a cone constraint and encode priorities as non-linear constraints, but this can be more computationally expensive. Lastly, a prioritized operational space formulation for gaze control was presented in \cite{SentisPhd}, however it is limited to the control of head gaze only, and the joint limit task insertion suffers from the same discontinuity issues mentioned previously while also not having a method for escaping the joint limit attractor.

\section{Technical Approach}
\subsection{Robot Kinematics and Jacobian}
\begin{figure}
\centerline{\includegraphics[width=0.9\columnwidth]{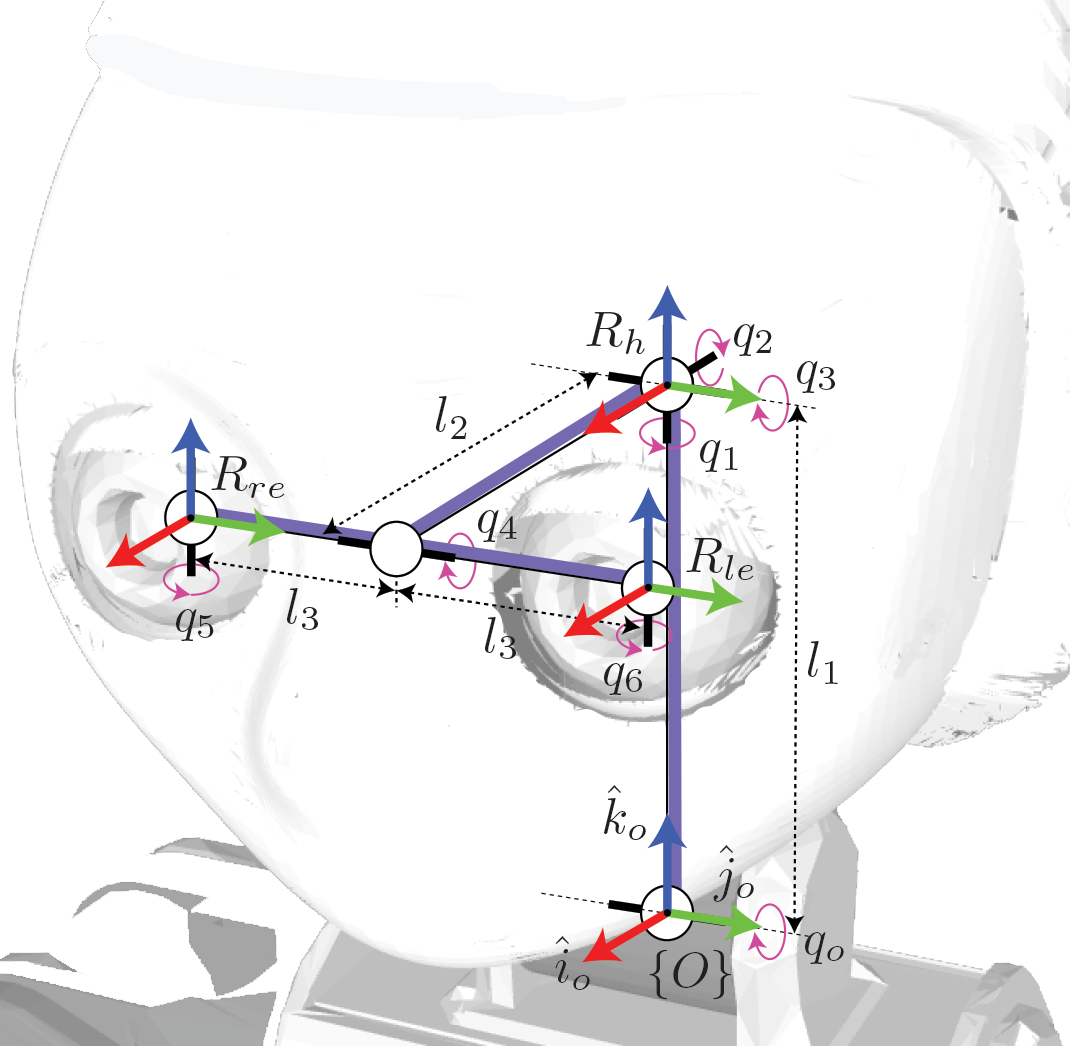}}
\caption{Dreamer has 7 Degree of Freedoms in its head. The 6-D spatial Jacobians is derived by first finding the Screw Axes of the kinematic chain (see Table \ref{t:dreamer_screw_axes}) and then recursively using the adjoint mapping operator. This methodology for deriving the Jacobian is explicitly described in \cite{modernrobotics}}
\label{fig:dreamer_kinematics}
\end{figure}


\begin{table}
\begin{center}
\caption{\textit{Dreamer Head Screw Axes} $S_i = (w_i, v_i)$ \textit{with link lengths} $(l_1, l_2, l_3) = (0.13849m, 0.12508m, 0.053m)$}\label{t:dreamer_screw_axes}
\begin{tabular}{ccc}
$S_i$ & $w_i \in \mathbb{R}^3$ & $v_i$ \\ 
 \toprule 
 $S_0$ & $(0, -1, 0)$ & $(0,0,0)$  \\ 
 $S_1$ & $(0,  0, 1)$ & $(0,0,0)$  \\
 $S_2$ & $(-1, 0, 0)$ & $(0,-l_1,0)$ \\
 $S_3$ & $(0, -1, 0)$ & $(l_1,0,0)$ \\
 $S_4$ & $(0, -1, 0)$ & $(l_1,0,-l_2)$ \\
 $S_5$ & $(0, 0, 1)$ & $(-l_3,-l_2,0)$ \\
 $S_6$ & $(0, 0, 1)$ & $(l_3,-l_2,0)$ 
\end{tabular}
\end{center}
\end{table}

The kinematics of Dreamer's head is described by Fig.~\ref{fig:dreamer_kinematics}. Let $q_0, q_1, q_2, q_3$ be the head joints, $q_4$ be the eye pitch joint and $q_5, q_6$ be the yaw joints for the right and left eyes respectively. 

Given an operational point $x \in \mathbb{R}^6$ on the robot's body with linear and rotational components, the spatial change, $dx$, with respect to the world frame due to a joint change $dq$ is described by 
\begin{align}
dx = J^s(q) dq,
\end{align}
where $J^s(q) \in \mathbb{R}^{6xn}$, is the 6-D spatial Jacobian of a robot with $n$ joints. Deriving $J^s(q)$ can be performed by first finding the screw axes of the kinematic chain (see Table \ref{t:dreamer_screw_axes}), and then recursively finding the $i$-th column, $J^s_{i}(q)$, of $J^s(q)$ using the adjoint mapping operator (See Ch.3 and Ch. 4 of \cite{modernrobotics}). Note that $J^s_{i}(q)$ describes the spatial twist as a function of the first $i$ joints $q_0, q_1, ..., q_{i}$. 

Setting $q = [q_0, q_1, ..., q_6]$, the spatial Jacobians of interest are
\begin{align}
J_{h}  &= [J^s_{0}, J^s_{1},  J^s_{2}, J^s_{3}, \ 0, \ \ \ 0, \ \ \ 0 \ ], \\
J_{re} &= [J^s_{0}, J^s_{1}, J^s_{2}, J^s_{3}, J^s_{4}, J^s_{5}, \ 0 \ ], \\
J_{le} &= [J^s_{0}, J^s_{1}, J^s_{2}, J^s_{3}, J^s_{4}, \ 0 \ , J^s_{6}], 
\end{align}
where the subscripts $h$, $re$, $le$ indicate the head, right eye, and left eye respectively. As it is trivial to control the operational space directions $[dx,dy,dz]$, here the focus is only on controlling the rotational components,  $[dw_x, dw_y, dw_z]$, of the operational space corresponding to roll, pitch, and yaw. Thus, for the Jacobian of the head, $J_{h}$ only the first three rows corresponding to head roll, pitch, and yaw. For the Jacobian of the eyes, $J_{le}$ and $J_{re}$, only the first two rows are kept to control eye pitch and yaw.

\subsection{Defining the Instantaneous Desired Gaze Orientation given a Fixation Point}
\begin{figure}
\centerline{\includegraphics[width=0.9\columnwidth]{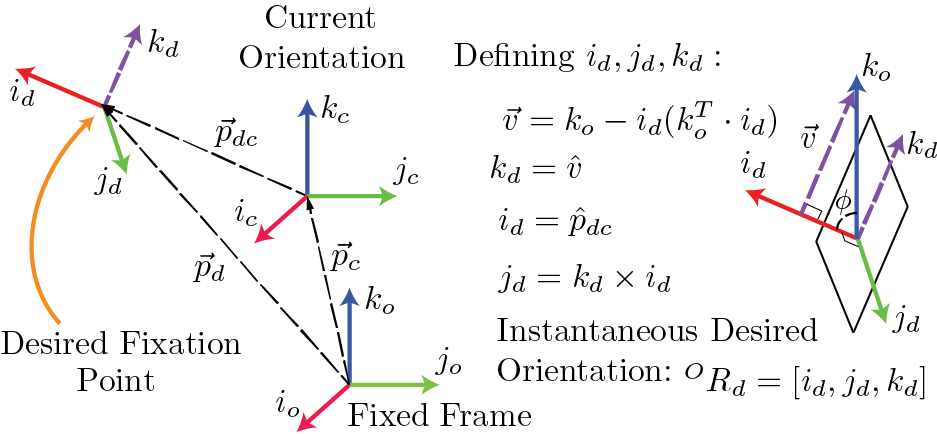}}
\caption{The Instantaneous Desired Gaze Orientation. An orientation, $^{O}R_d$ w.r.t to the world frame is constructed using the current orientation of the operational space frame and the desired fixation point. The vector $\vec{p}_{dc}$ defines the desired unit vector direction, $i_d$, which also defines the normal of a plane. A unit vector (here $k_o$ is used) from the fixed frame is then used to construct $k_d$ by projecting $k_o$ to the plane defined by $i_d$. Finally, $j_d$ is constructed by taking the cross product of $k_d$ and $i_d$ }
\label{fig:inst_desired_frame}
\end{figure}
The control structure presented here constantly steers the current head and eye orientations to point towards the corresponding desired fixation points. At every time step, an instantaneous desired gaze orientation is constructed. 

Note that a rotation matrix, $R = [i, j, k] \in \mathbb{R}^{3x3}$, with unit vector columns,  $i,j,k \in \mathbb{R}^3$, can be used to represent the orientation of a frame with respect to (w.r.t) a  reference frame. Thus, defining the instantaneous desired gaze orientation is equivalent to finding the  instantaneous desired unit vectors. Let $i$, $j$, and  $k$ be unit vectors and the subscripts $o$, $c$, and $d$ indicate the names world, current and desired orientations respectively. All the unit vectors are w.r.t to the world frame. Next let $\vec{p}_d$ and $\vec{p}_c$ be the location of the fixation point and the origin of the operational space frame. Finally, let $\vec{p}_{dc} = \vec{p}_d - \vec{p}_c$. Using Figure~\ref{fig:inst_desired_frame} as a visual reference, we obtain
\begin{align}
i_d &= \frac{\vec{p}_{dc}}{||\vec{p}_{dc}||} = \hat{p}_{dc}, \label{eq:inst_eq_start} \\
v &= f_o - i_d(f^T_o \cdot i_d), \\
k_d &= \frac{\vec{v}}{||v||} = \hat{v}, \\
j_d &= k_d \times i_d \label{eq:inst_eq_end},
\end{align}
where $f_o$ is a fixed frame unit vector. Therefore the instantenous desired orientation is $^{O}R_d = [i_d, j_d, k_d]$. The choice of $f_o$ depends on user need and the desired generated behavior. In our case, the unit vector $i_c$ and fixation points have positive world frame x-coordinates so $f_o$ is selected to be $k_o$.

\subsection{Defining the World Frame Orientation Error}
Let $^{O}R_c$ and $^{O}R_d$ be the rotation matrices w.r.t frame {O} describing the robot's current and desired end-effector orientation frames respectively . The goal is to find the rotation matrix described in the world frame that will bring $^{O}R_c$ to $^{O}R_d$. 


Remembering that pre-multiplying a reference frame, $^{O}R_A$ (described as a rotation matrix) by a rotation matrix $^{O}R_B$ results to an extrinsic rotation  of frame $^{O}R_A$ by $^{O}R_B$ in frame ${O}$. the rotation matrix which will rotate frame $^{O}R_c$ to $^{O}R_d$ in the world frame is referred to as the orientation error\footnote{This is equivalent to finding the total rotation performed by SLERP} matrix, $^{O}R_e$.  It can be solved via
\begin{align}
^{O}R_e \ ^{O}R_c &= ^{O}R_d \\
^{O}R_e &\triangleq ^{O}R_d (^{O}R_c)^{-1} \label{eq:rotation_error_R}.
\end{align}

Next, this rotational frame error is described in terms of quaternions. The reader is referred to the appendix of \cite{modernrobotics} for a primer on unit quaternions. The unit quaternion with respect to frame $O$ is defined to be
\begin{align}
^{O}q^t = [\cos(\theta/2), \ \hat{\omega} \ \sin(\theta/2)] \in \mathbb{R}^4 ,
\end{align}
where $\theta \in [0, \pi]$ is the right-hand rotation about a unit vector axis, $\hat{\omega} \in \mathbb{R}^3$ rotation . Note that $\theta$ and $\hat{\omega}$ are the axis-angle representation of the quaternion.

Given a rotation matrix $R$, the elements of its corresponding unit quaternions, $\pm q^t$ can be obtained. For consistency, the unit quaternion, when converted to its axis-angle representation, with an angle $\theta \in [0, \pi]$ is always selected. Then the quaternion error, $^{O}q^t_e$ is 
\begin{align}
^{O}q^t_e = ^{O}q^t_d \otimes {^{O}q^{t}_{r}}^{-1}, \label{eq:rotation_error_q}
\end{align}
where the inverse of the unit quaternion is simply the $-\hat{\omega}$ of its axis angle-representation, and the operator $\otimes$ is the unit-quaternion product.  

\subsection{The Operational Space Task For Orientation Control}\label{section:orientation_error_operational_task}
Having specified the orientation error $^{O}q^t_e$ , the operational space task can now be specified which will bring the current orientation $^{O}q^t_c$ to a desired orientation $^{O}q^t_d$. 

To do so, we note that the quaternion error derived earlier is with respect to the world frame and that a quaternion can be decomposed into its axis-angle components, $\theta$ and $\hat{\omega}$. Specifically, for $^{O}q^t_e$, the product of its axis-angle representation,  $\hat{\omega}_e \theta_e$, is equivalent to the angular velocity needed in one second to rotate frame $^{O}q^t_r$ to $^{O}q^t_d$. 
For small $dt$ the operational orientation task steers $q^t_c$ towards $q^t_d$ by defining $dx$ as
\begin{align}
dx &= k \hat{\omega_e} \theta_e  \in \mathbb{R}^3, \label{eq:desired_task}
\end{align}
with an appropriate operational task gain $k$. Here, $k=1$.

\subsection{Orientation Control}
\label{section:orientation_control}
\begin{figure*}
  \centerline{\includegraphics[width=0.9\textwidth]{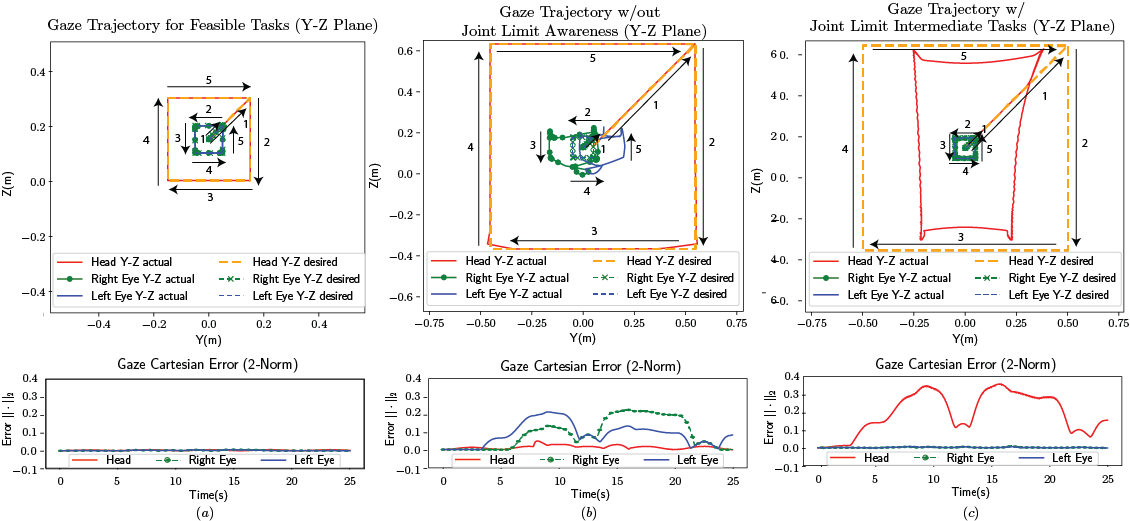}}
  \caption{The gaze trajectory of a prioritized controller with no joint limit tasks (a and b), and  with intermediate joint limit tasks (c). The task is to trace a small square for the eyes and a bigger square for the head the arrows indicate the waypoint  trajectory order. The eye tasks have higher priority than the head. In (a), the  the gaze trajectory tasks for the head and eyes are within joint limits so there is perfect gaze tracking.  In (b), since joint limits are not part of the controller constraints, tracking for the eye tasks fail as the robot continues to generate $dq$ commands in the eye joints. In (c) the low priority head task is executed without disturbing the satisfaction of the higher priority eye tasks.}  \label{fig:controller_discussion}
\end{figure*}

\subsubsection{Head Orientation Control as a Single Task}
For robot heads without eyes, only a single fixation point orientation task is needed. The following resolved motion rate control \cite{liegeois1977automatic} with our operational space definition for $dx$ is enough:
\begin{align}
\label{eq:resolved_motion_control_1}
dq &= J^{\dagger}dx,
\end{align}
where $J^{\dagger}$ is the Moore-Penrose pseudo inverse of the Jacobian. However, head-eye robots naturally have two fixation points, one for the head and the other for the eyes. 

\subsubsection{Simultaneous Head-Eye Orientation Control as Separable Tasks}
Note that for head-eye robots, the orientation tasks for the head and the eyes are separable as the head and eyes each have enough degrees of freedom to control the head-eye robot towards multiple feasible fixation points. Where \textit{feasible} here means that the fixation point is within the joint limits of the robot. In other words, the eye degrees of freedom and the head degrees of freedom are independently coordinated to point at different fixation points.

Concretely, this can be done by constructing the spatial Jacobians for the head and eyes as $J^s_{h}$ and $J^s_{e}$ with zero columns that correspond to eye and head joints respectively
\begin{align}
J^s_{h}  &= [J_{h}(q) , 0 ], \\
J^s_{e} &= [0, J_{e}(q)],
\end{align}
where $J_{h}(q)$ is the spatial Jacobian with head joints only and $J_{e}(q)$ is the spatial Jacobian with eye joints only. With our definition of operational space tasks $dx_{h}$ and $dx_{e}$, stacking them such that $J = [J^T_{h}, J^T_{e}]^T$ and $dx = [dx^T_{h}, dx^T_{e}]^T$ and using Eq(\ref{eq:resolved_motion_control_1}) will control the head-eye robot towards the fixation point. 

However, this approach has two significant limitations. It has no notion of joint limits or prioritization. Under eye joint limits, if a user cares more about the eye fixation point over the head's fixation point, the user must analyze if the gaze fixation point for the eye is reachable given the current head configuration. It will be more desirable to first satisfy the eye fixation point (priority 1) and then attempt to satisfy the head fixation point (priority 2) .

\subsubsection{Simultaneous Head-Eye Orientation Control with Priorities}
To enforce priorities for operational tasks $x_1$ and $x_2$, the following control structure may be implemented.
\begin{align}
dq &= dq_1 + dq_2, \\ \label{eq:standard_priority}
dq_1 &= J_1 ^\dagger dx_1, \\
dq_2 &= (J_2 N_1)^\dagger (dx_2 - J_2 dq_1),
\end{align}
where $N_1 = I - J_1^\dagger J_1$ is the null space projector due to task 1. The reader is reffered to \cite{baerlocher1998task,slotine1991general} for a review on setting up kinematic prioritized tasks and its recursive formulation.

While this approach has prioritization, it still has no notion of joint limits. This control approach is implemented in Fig.~\ref{fig:controller_discussion}(a) and (b). where the eye gaze task has higher priority than the head gaze task. Since the formulation has no notion of joint limits, when the eye yaw joint limits are hit, the controller Eq(~\ref{eq:standard_priority}) continues to generate $dq$'s for the eye joints (See Fig~\ref{fig:controller_discussion}b). If the eyes or the task specification hits no joint limits, this formulation will be correct (See Fig~\ref{fig:controller_discussion}a). 



\subsubsection{Simultaneous Head-Eye Orientation Control with Joint Limits and Priorities}

To address the limitations of the above controller, we introduce a task hierarchical framework with joint limits.


Let $n$ be the number of robot joints, $m$ be the number of joints that have limits and $t$ be the number of operational tasks. Our prioritized controller will have $t+1$ prioritized tasks with the joint limit tasks having the highest priority. Each joint limit task must have a task Jacobian defined as a row vector,
\begin{align}
J_{1,j} = [0, ..., 0, 1, 0, ..., 0] \in \mathbb{R}^{1 \times n}
\end{align}
where the position of $1$ is the column corresponding to the joint. The joint limit task Jacobian, $J_1$ is expressed by stacking $J_{1,j}$ as $J_1 = [J^T_{1,1}, J^T_{1,2}, ... ,J^T_{1,j}, ...,J^T_{1,m}]^T$. The lower priority tasks $2,...,t+1$ will be the task Jacobians for the eye and head gaze tasks. Here, the prioritized gaze fixation point tasks $2$ and $3$ will be the eye and head orientation tasks respectively.

However, each joint limit task should only activate when the joint enters an activation buffer. We utilize the intermediate task transition formulation for smooth task transitions \cite{lee2012intermediate}. The control structure for this formulation is as follows:
\begin{align}
dq &= dq_1 + dq_2 + dq_3 + ... + dq_{t+1}, \label{eq:original_dq_sol} \\
dq_1 &= J_1 ^\dagger dx^i_1, \\
dq_2 &= (J_2 N_1)^\dagger (dx_2 - J_2 dq_1), \\
dq_3 &= (J_3 N_1 N_{2|1})^\dagger (dx_3 - J_3 (dq_1 + dq_2)), \\
&\cdots, \nonumber \\
dq_{t+1} &= (J_{t+1} N_{[t]})^\dagger  (dx_{t+1} - J_{t+1} \sum_{k=1}^{t} dq_k), \label{eq:dq_sol_end} \nonumber \\
\end{align}
where $dx^i_1 = [dx^i_{1,1}, ..., dx^i_{1,j}, ..., dx^i_{1,m}]^T$ is the desired intermediate value for the joint limit tasks $j \in \{1,2,...,m\}$, defined below, $J_s$ is the $s$-th task Jacobian, defined previously, $N_{[k]}$ is the nullspace projector due to the higher priority tasks $k, k-1, ... ,1$, defined as,
\begin{align}
\label{eq:task_nullspace}
N_{[k]} = \prod_{s=1}^{k} N_{s|s-1}, 
\end{align}
and  $N_{s|s-1}$ is also a nullspace projector due to tasks $s, s-1, ..., 1$, recursively defined as
\begin{align}
\label{eq:nullspace_proj}
N_{s|s-1} &= I - (J_s N_{s-1|s-2})^\dagger(J_s N_{s-1|s-2}) \\
&\textmd{ Base Case: } N_{1|0} = N_1 = (I - J_1^\dagger J_1). \nonumber
\end{align}

Here, a special case of the IDV is used in which the only intermediate tasks are due to joint limits. Thus, only the joint limit tasks, $dx^i_1$, needs to be computed recursively. The $j$-th joint limit task is computed as
\begin{align}
dx^i_{1j} = h_j(q) dx_{1,j} + (1-h_j(q)) J_{1,j} dq_{[\setminus j]},
\end{align}
where $dx_{1,j}$ is the usual desired task value for the joint limit, $h_j(q) \in [0,1]$ is the task activation parameter due to a joint configuration $q$, and $dq_{[\setminus j]}$ is the solution without the joint limit task $j$. Since only joint tasks will activate ($h_j(q) = 1$) or deactivate ($h_j(q) = 0$), $h_j(q)$ is the same activation function defined in \cite{lee2012intermediate}. Instead of permanently attracting the joint limit task \cite{sentis2005control},
it is desirable that the joint attempts to leave the activation buffer so that the robot can regain the degree of freedom. Thus, the desired values for the joint limit avoidance task is 
\begin{align}
dx_{1,j} = k_j(\mu_{q_j} - q_j), \textmd{ for joint } j \in {1,...,m}
\end{align}
where $\mu_{q_j}$ is the center of the joint, and $k_j$ is an appropriate gain (set to $k=0.001$), which will bring the joint away from the activation buffers.

Finally, we define $dq_{[\setminus j]}$, the task solution without the joint limit task $j$. Concretely, $dq_{[\setminus j]}$ calls another instantiation of Eqs.(\ref{eq:original_dq_sol} - \ref{eq:dq_sol_end}) but without the $J_{1,j}$ row in the joint limit task Jacobian of $J_1$. At each call, a row of $J_1$ is removed. As joint limits are the only intermediate values considered here, the base case for $dq_{[\setminus j]}$ is the regular prioritized solutions without any joint limit task. A pseudocode of the algorithm in python notation is provided in Algorithm \ref{alg:prio_controller_idv}.



\subsection{Minimum Jerk Trajectory Generation and Tracking}
For gaze behavior generation, the controller and the task error definition described in Sec.\ref{section:orientation_error_operational_task} can be used to follow trajectories designed in Cartesian space. Concretely, Cartesian trajectories can be constructed from the current gaze fixation point $\vec{x}_o \in \mathbb{R}^3$, to a final point $\vec{x}_f \in \mathbb{R}^3$. A minimum jerk trajectory \cite{flash1985coordination} for each Cartesian dimension $(^Ox,^Oy,^Oz)$ in the fixed frame $O$ can be constructed using a 5-th order polynomial, $s(t)$ defined below, with boundary conditions on the position, velocity, and acceleration described as a vector $b = [s(t_i), s(t_f), \dot{s}(t_i), \dot{s}(t_f), \ddot{s}(t_i), \ddot{s}(t_f)]^T$, where $t_i$ and $t_f$ indicate initial and final times respectively.
\begin{align}
s(t) = a_o + a_1 t + a_2 t_2^2 + a_3 t_3^3 + a_4 t_4^4 + a_5 t_5^5.
\end{align}
For a single dimension, finding the coefficients $\vec{a} = [a_o, ..., a_5]^T$ can be done by solving for $\vec{a}$ in $B\vec{a} = \vec{b}$, where $B \in \mathbb{R}^{6\times6}$ is the corresponding matrix with $t_i$ and $t_f$ terms.

To perform gaze tracking on a given Cartesian trajectory, $\vec{s}(t) = [s_x(t), s_y(t), s_z(t)]^T$, at each time $t$, the instantaneous desired orientation is constructed by using Eqs.(\ref{eq:inst_eq_start}-\ref{eq:inst_eq_end}) and setting $\vec{p}_{d} = \vec{s}(t)$. This generates the instantaneous desired orientation $^OR_d$. Then the rotation error can be extracted with Eqs.(\ref{eq:rotation_error_R}) and (\ref{eq:rotation_error_q}) and the operational space task $dx$ at this time step is extracted with Eq.(\ref{eq:desired_task}). This $dx$ is the input to the operational space controller in Sec. \ref{section:orientation_control}.



\begin{algorithm}
\caption{Recursive Formulation of a Prioritized Controller with Intermediate Values}\label{alg:prio_controller_idv}
\begin{algorithmic}[1]
\State{\text{// Initialize $m$ joint limit tasks}}
\State {$\ (J_j,dx_{j}) = ([J_{j_1},..., J_{j_m}],[dx_{j_1}, dx_{j_2}, ...., dx_{j_m}])$}
\State {$\ h = [h_{1}, h_{2}, ...., h_{m}]$}

\State{\text{// Initialize $t$ operational space tasks }}
\State {\ $(J_{o},dx_{o}) = ([J_{o_1}, ..., J_{o_t}],  [dx_{o_1}, ..., dx_{o_t}])$ }

\Procedure{$dq$($J_j$, $dx_{j}$, $h$,$J_{o}$,$dx_{o}$ ):}{}

\Procedure{$dq_{[\setminus j]}(j)$:}{}
\State{\text{// Compute $dq$ w/out joint limit task $j$}}
\State $J_{j{[\setminus j]}} = J_j[:j] + J_j[j+1:]$
\State $dx_{j{[\setminus j]}} = dx_{j}[:j] + dx_{j}[j+1:]$
\State $h_{[\setminus j]} = h[:j] + h[j+1:]$
\State \Return{$dq( J_{j_{[\setminus j]}}, dx_{j_{[\setminus j]}}, h_{[\setminus j]}, J_o, dx_o ) $ }
\EndProcedure

\State $|h| = \text{length of } h$
\State $(J_T, dx_T) = (J_o, dx_o)$

\If{$|h|>0$}
\State{\text{// Stack the joint limit constraints} }
\State{$J_{1} = [J^T_{j_1}, J^T_{j_2}, ..., J^T_{j_{|h|}}]^T$ }

\For{$j=1$ \textit{ to } $|h|$}
\State{\text{// Compute IDV due to joint $j$}}
\State{\text{// Note the recursive call to $dq_{[\setminus j]}$}}
\State{$dx^i_j = h[j]dx_{j}[j] + (1-h[j])J_j[j] dq_{[\setminus j]}(j)$}
\EndFor

\State{$J_T = [J_{1}] + J_T$ \text{//combine lists}}
\State{$dx_T = [dx^i_1, ..., dx^i_{|h|}] + dx_o$ \text{//combine lists}}
\EndIf
\State $|t| = \text{length of } J_T$
\State {\text{//Pre-compute $N_{[1]}, N_{[2]}, ..., N_{[t-1|]}$ given $J_T$}}
\State $dq_{\Sigma} = 0 \in \mathbb{R}^{n}$ 

\For{$k = 1$ \textit{ to } $|t|$}
\If{$k = 1$}
\State {$dq_k = (J_T[k])^\dagger dx_T[k]$}
\Else
\State  {$dq_k = (J_T[k] N_{[k-1]} )^\dagger (dx_T[k] + J_T[k] dq_{\Sigma})$}
\EndIf
\State {$dq_{\Sigma} = dq_{\Sigma} + dq_k$ \text{// task $k$ contribution to $dq$}}

\EndFor
\State \Return{$dq_{\Sigma}$}

\EndProcedure
\end{algorithmic}
\end{algorithm}

\section{Controller Experiments and Results}
\begin{figure*}
  \centerline{\includegraphics[width=0.9\textwidth]{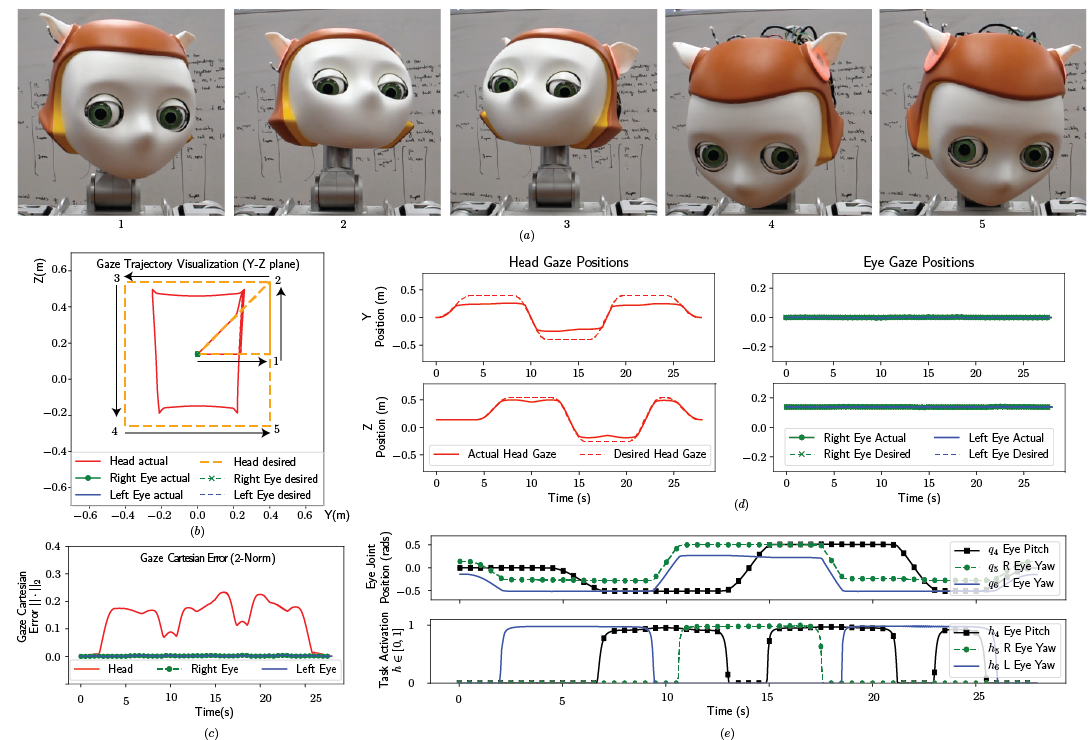}}
  \caption{The higher priority eye gaze tasks are commanded to look at a fixed point and the lower prioritized head gaze task is commanded to trace a square, which will cause the eye joint limits to hit during task execution. (a) Shows the head-eye configuration at the specified waypoints. (b) Shows the actual trajectory of the head and eye gaze tasks. The dotted-orange and solid red lines indicate the desired and actual head gaze trajectory respectively following task priority constraints. The desired and actual eye gaze positions remain at the fixation point. (c) Shows that gaze Cartesian error is only present on the head gaze task (d) shows the minimum-jerk based trajectories for the head with perfect fixation point tracking for the eyes. (e) Shows the eye joint positions and the corresponding $h$ joint limit activation values.}  \label{fig:dreamer_executing_fixed_point_eye_task}
\end{figure*}

Controller validation is performed on the real Dreamer robot as shown in Fig.\ref{fig:dreamer_executing_fixed_point_eye_task}. The robot is tasked with three orientation trajectories, two for the eyes, and one for the head with the eye gaze tasks having higher priority than the head gaze task. The eye is commanded to stay fixated at a 3D point directly in front of the robot, while the head is commanded to create a square by following way points defining a minimum jerk trajectory. While both tasks cannot be accomplished simultaneously, the controller must maintain the eye fixation point and only execute the lower priority head gaze task if it can be done without interference.

As Fig.\ref{fig:dreamer_executing_fixed_point_eye_task} shows, our controller preserves task prioritization even under joint limits (Fig.\ref{fig:dreamer_executing_fixed_point_eye_task}. a and e). Notice that only the head gaze task has a Cartesian 2-norm error (Fig.\ref{fig:dreamer_executing_fixed_point_eye_task}c) and the eye gaze Cartesian positions are tracked perfectly (Fig.\ref{fig:dreamer_executing_fixed_point_eye_task}d). Finally, the joint limit avoidance tasks are continuously inserted and removed, with the corresponding $h$ activation values, as the eye joints approach their limits (Fig.\ref{fig:dreamer_executing_fixed_point_eye_task}e). Due to task prioritization with joint-limit awareness, the controller maintains the gaze fixation task. Note that this biomimetic behavior of the vestibulo-occular-reflex (VOR) \cite{fetter2007vestibulo} naturally occurs in our controller.

\section{Discussion and Conclusions}
Inspired from projection-based whole-body controllers, a generic controller with task prioritization for joint-constrained head-eye robots is presented and experimentally validated on the Dreamer humanoid robot.  In order to formulate simultaneous gaze tasks as operational space inputs to the controller, the construction of the instantaneous desired orientation was presented. To handle intermediate joint limits without solution discontinuity, the IDV approach is utilized and described in detail with an accompanying pseudo code. Finally gaze behavior is generated via gaze tracking of minimum jerk trajectories in Cartesian space.

The Cartesian specification of gaze trajectories transforms the problem of 
trajectory generation in joint space to Cartesian space, which has lower dimensions. As a future work, emotive behavior generation using Cartesian space trajectories may enable skill transfer of head-eye behavior, such as expressing different emotions, across many robots.

To conclude, the presented head-eye controller addresses the missing capability of handling multiple 3D gaze tasks with priorities under joint limits. This generic controller can enable users to execute precise gaze control for enhancing human-robot-interactions. 






\bibliographystyle{IEEEtran}
\bibliography{IEEEabrv,IEEEexample,my_bib}


\end{document}